\documentclass[10pt,letterpaper]{article}
\usepackage[top=0.85in,left=2.75in,footskip=0.75in,marginparwidth=2in]{geometry}

\usepackage[utf8]{inputenc}

\usepackage{cite}

\usepackage{nameref,hyperref}

\usepackage{microtype}
\DisableLigatures[f]{encoding = *, family = * }

\raggedright
\usepackage{changepage}

\usepackage[aboveskip=1pt,labelfont=bf,labelsep=period,singlelinecheck=off]{caption}

\makeatletter
\renewcommand{\@biblabel}[1]{\quad#1.}
\makeatother

\usepackage{lastpage,fancyhdr,graphicx}
\usepackage{epstopdf}
\fancyheadoffset[L]{2.25in}
\fancyfootoffset[L]{2.25in}

\usepackage{color}

\definecolor{Gray}{gray}{.25}

\usepackage{graphicx}

\usepackage{sidecap}

\usepackage{wrapfig}
\usepackage[pscoord]{eso-pic}
\usepackage[fulladjust]{marginnote}
\reversemarginpar

\begin{document}
\vspace*{0.35in}

\begin{flushleft}
{\Large
\textbf\newline{Machine learning approaches for identifying prey handling activity in otariid pinnipeds}
}
\newline
\\
Rita Pucci\textsuperscript{1,*},
Alessio Micheli\textsuperscript{2},
Stefano Chessa\textsuperscript{2},
Jane Hunter\textsuperscript{3},
\\
\bigskip
\bf{1} University of Udine
\\
\bf{2} University of Pisa
\\
\bf{3} The University of Queensland
\\
\bigskip
* rita.pucci@uniud.it

\end{flushleft}

\section*{Abstract}
Systems developed in wearable devices with sensors on board are widely used to collect data of  of humans and animals activities with the prospective of an on-board automatic classification of data. An interesting application of these systems is to support animals' behaviour monitoring gathered by sensors' data analysis. This is a challenging area and in particular with fixed memories capabilities because the devices should be able to operate autonomously for long periods before being retrieved by human operators, and being able to classify activities on board can significantly improve their autonomy. In this paper we focus on the identification of prey handling activity in seals (when the animal start attaching and biting the prey), which is one of the main movement that identify a successful foraging activity. Data taken into consideration are streams of 3D accelerometers and depth sensors values collected by devices attached directly on seals. To analyse these data, we propose an automatic model based on Machine Learning (ML) algorithms. In particular we compare the performance (in terms of accuracy and F1score) of three ML algorithms: Input Delay Neural Networks, Support Vector Machines, and Echo State Networks. We attend to the final aim of developing an automatic classifier on-board. For this purpose, in this paper the comparison is performed concerning the performance obtained by each ML approach developed and its memory footprint. In the end, we highlight the advantage of using an ML algorithm, in terms of feasibility in wild animals' monitoring.


\section*{Introduction}
Scientific publications about ''bio-logging'', the attachment of simple recording devices to animals to collect information, first appeared in the 1950s. The term bio-logging was first introduced by Ropert-Coudert and Wilson in \cite{ropert2004}, who distinguished it from biotelemetry \cite{boyd2004bio}. Bio-logging means to log signals collected by embedded sensors onto a device attached to an animal. Biotelemetry, on the other hand, involves the transmission of signals, collected through sensors attached to the animal, via a wireless connection. The bio-logging technique allows the collection of data even when the device (i.e., the animal) is outside of the wireless range for transmitting to the remote data collection station but it is necessary to physically retrieved the device to access to the data. These days, hybrid systems that combine bio-logging and biotelemetry provide a solution to this issue. Bio-logging is widely applied in biology, and in particular in ecology for investigating animal behaviour via analysis of an individual's activities. For example, the information provided by sensors enables ecologists to observe how an animal (e.g., a seal) manages its body energy \cite{gallon2013identifying} or how it changes its behaviour in relation to prey encounter rate \cite{LeBras2016seals}. 

\noindent Bio-logging methods aimed at the identification of individual activities typically involve cameras, internal sensors and/or external sensors \cite{naito2004bio}.  For short monitoring periods, video recordings provide direct evidence of the activity performed by the animal. Videos are recently used to estimate to pose of dogs, by a multidisciplinary system that can identify the flanks and paws of the animal, in \cite{haggag2016semantic}. Examples of the use of video cameras for  bio-logging can be found in \cite{davis2008}, \cite{moll2007new}, and \cite{volpov2015identification}. However, for longer monitoring periods, video recordings demand a significant amount of digital memory that is not usually available on such animal tagging devices and the researchers need to retrieve the video camera and analyse the recorded video content to identify activities.

\noindent Internal sensors provide accurate physiological information for inferring the health and the activity of the animal \cite{wilson1992short}, \cite{kuhn2009time}, \cite{ancel1997prey}. Although the recorded data requires a relatively low amount of memory for storage, despite in this case retrieving the sensors can be challenging.

\noindent External sensors have an advantage compared to the internal sensors with regard to retrieval. With external sensors, the information is not physiological but physical (speed, acceleration, magnetic fields, and external pressure). The use of sensors such as accelerometers for bio-logging is widespread in the literature, as evidenced by \cite{dubois2009thermoregulation}, \cite{miyashita2014}, \cite{barbuti2016}, \cite{gao2013web}, and \cite{soltisWilson}. Internal sensors, such as stomach and oesophagus temperature sensors, and simpler external sensors, such as accelerometers, magnetometers, and pressure sensors, are valid alternatives to video cameras. Video monitoring requires memory in order of gigabytes, whilst accelerometers require memory in order of megabytes. Furthermore, poses and activities classification can be done both with video recording and external sensors, and it is worth noting that external sensors are more reliable under different environmental conditions. Unlike video cameras, sensors such as accelerometers, magnetometers, and pressure sensors are not dependent on weather or light conditions. Moreover by using sensors, we cope with data collected by sensors for activities recognition adding the challenge of interpreting noise and unprocessed data.

\noindent In this paper, we focus our attention on bio-logging to identify prey handling behaviour in seals (\textit{otariid pinnipeds}), using accelerometer and pressure (depth) data. Seals' prey-handling behaviour provides valuable information about their ecology but it is extremely difficult to monitor because it needs to be  observed underwater, at great depth, in extreme cold and over vast areas  \cite{Naito2007101},\cite{Naito2010309}, \cite{suzuki2009validation}, \cite{Viviant2014}, \cite{ gallon2013identifying}, and \cite{skinner2009head}. The aim is to develop and evaluate an automatic classifier able to identify the prey handling activities from accelerometer data. We develop the classified in order to be installed and to perform classification on board of the tagging device, the focus on an on-board system reduce the demand of memory space by selecting the information kept on the device. Hence, the classifier design needs to take into consideration the limited memory and processing capabilities available on board the device. The classifier design has to balance an optimum trade-off between classification accuracy and speed, memory and processing capabilities, and the size and weight of the device.

\noindent ML approaches \cite{hastie2009elements} have been widely employed for recognising human activities, such as \cite{kim2010human}, \cite{lara2013survey}, \cite{palumbo2016human}, and \cite{hu2004learning}. In this paper, we explore such approaches for analysing accelerometer data from captive otariid pinnipeds (seals). We select three different ML approaches with different characteristics with respect to the proposed issues: Input Delay Neural Network (IDNN)\cite{Haykin2009}, Support Vector Machine (SVM)\cite{vapnik1998statistical}, and Echo State Network (ESN)\cite{gallicchio2011architectural} (which are all of the three naturally applied to pattern recognition). The former approaches have been applied with bio$-$logging data from both humans and animals. The latter has been successfully applied to analyse human movements \cite{Amato2012Rubicon}, \cite{Bacciu2013Rubicon} and here applied for the first time over bio$-$logging data of animals. More specifically, we compare the performance and the feasibility of these three approaches by analysing accelerometer data streams collected from seals at Taronga Zoo in Sydney by a research team from Macquarie University \cite{Ladds2016seals}. In \cite{Ladds2016seals}, authors compare different data analysis approaches by mean the performance without analysing the feasibility of the approach on$-$board of the device. Furthermore, we evaluate the generalisability of the approaches, where with term generalisability we consider the pre-processing necessary for the application of a specific approach.

\noindent In the following sections, we describe and evaluate an embedded classifier that we have developed to automatically identify prey handling activities in seals (but which is applicable to activity recognition over a wide range of species). We discuss how the outcomes will reduce the magnitude of output data and reduce the frequency of device retrieval. Deployment of such as classifier on emerging hybrid devices augmented with biotelemetry that uses low rate, long range wireless networks (such as \cite{vangelista2015long}) will enable near-real time upload of animal activity summaries to remote data collection stations. 

\section{Materials and Methods}
\subsection{Dataset}
The dataset used in this paper is an extract of an original dataset available from the PlosOne website at \cite{Ladds2016seals}. For the purpose of this paper, we chose data subsets that specifically include bouts of foraging for prey. In the dataset the subjects of interest are seven different seals species: two Australian fur seals (\textit{Arctocephalus pusillus deriferus}), three New Zealand fur seals (\textit{Arctocephalus forsteri}), one sub$-$antarctic fur seal (\textit{Arctocephalus tropicalis}), and two Australian sea lions (\textit{Neophoca cinerca}). They were observed in their natural environment respecting marine facilities. They were fed and cared by guidelines of individual facility. For details concerning the recording campaign it is to consider the Ladds et al. paper \cite{Ladds2016seals}.

\noindent The data was recorded using a 3D accelerometer embedded in a G6a and G6a+ data loggers, from CEFAS technology, attached to the back of seals. 

\begin{figure}[!ht]
    \centering
    \centerline{\includegraphics[width=7cm]{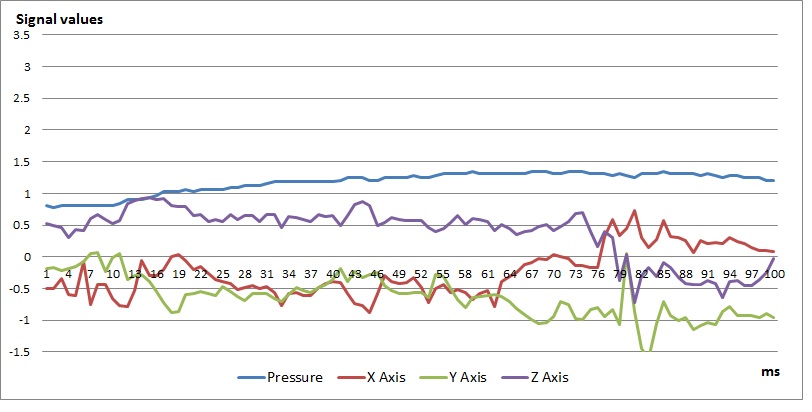}}
    \caption{An example of raw acceleration data and pressure data of a prey handling activity.}
    \label{SealPreyHandling}
\end{figure}

\begin{figure}[!ht]
    \centering
    \centerline{\includegraphics[width=7cm]{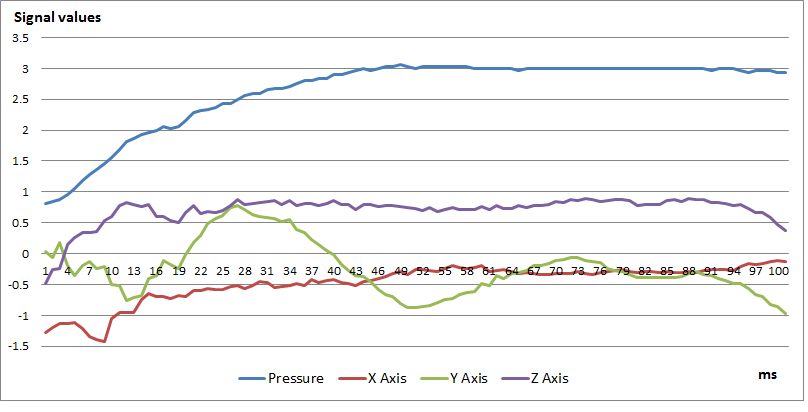}}
    \caption{An example of raw acceleration data and pressure data of a swimming (non$-$prey handling) activity.}
    \label{SealSwimming}
\end{figure}

\noindent The exact position of the device on the back of the animal depends to the species. For the Australian fur seals, the accelerometer was attached between shoulder blades on the top layer of fur using the Tesa tape, and for the sea lions and sub$-$antarctic fur seals it is used a customised harness. The harness is fixed with three clips, one behind the neck and two at the back, and a pocket containing the accelerometer is sewn into the harness. The data from the accelerometer records as time-series of 4$-$tuples movement along three spatial axes (heave, surge, and sway) and the depth. Each stream is recorded at 25 Hz sampling rate, the time laps is not homogeneous for all the recordings due to the duration of the swimming activity. Figures \ref{SealPreyHandling}, and \ref{SealSwimming}, illustrate a portion of the data streams recorded during a prey handling activity, and during a swimming (none-prey handling) activity, respectively.


\subsection{Methods}
\label{method}

Since we have time-series of different length, we apply two formats: (i) extracted windows of fixed length from time series, or (ii) windows with equal length time steps across the series. In the first format, each window is classified individually. Whilst in the second format, we classify each time step in the time series using the historical information from previous steps. In this paper, we compare the results of applying different classifiers to both formats, which we refer to as the ``\textit{windows}" and ``\textit{streams}" formats respectively.

\noindent For the ``windows" format, the time-series are firstly split in windows, with an overlap of half window. We take into consideration two different lengths: 1 second with overlap of 0.5 seconds and 0.4 seconds with overlap of 0.2 seconds. Since the resolution is 25 Hz, each window of 1 second has 100 values, 25 values for each accelerometer axes plus 25 for the pressure signal, and each window of 0.4 seconds has 40 values, 10 values for each accelerometer axes plus 10 values for the pressure signal. We obtained a dataset with 12692 balanced windows (6346 no-prey handling activity, and 6346 prey handling activity) of 1 second each, and a dataset with of 12732 balanced windows (6366 no-prey  handling  activity,  6366 prey  handling  activity) of 0.4 seconds each. Hereafter we use the following terms to refer to the four different datasets:
\begin{itemize}
\item Raw\_1, to the dataset of 1 second windows;
\item Features\_1, to the dataset of features extracted from 1 second windows;
\item Raw\_0.4, to the dataset of 0.4 second windows;
\item Features\_0.4, to the dataset of features extracted from 0.4 second windows.
\end{itemize}

\noindent With the ``streams" format, we consider the entire time-series till the last time step collected. In both formats each window or time step is labelled with 1 if it is classified as prey handling and with -1 otherwise. 

\noindent Based on the datasets we implemented the automatic classifier discriminating between the two formats. The automatic classifier for the ``windows" discerns between the analysis of features vectors from vectors of raw data. In the first case, the classifier format consists of two stages. The first stage is the filter stage, which involves performing feature extraction on each window. The features selected in this work are in accordance with  \cite{lara2013survey}, and \cite{Carroll2016} and consists of \textit{mean}, \textit{standard deviation}, \textit{minimum}, \textit{maximum}, \textit{skew}, \textit{kurtosis}, and \textit{pairwise correlation}. The features vector obtained comprises 30 values (7 feature values for three axes and the depth value) for both windows of 1 second and windows of 0.4 seconds. The second stage is the classifier and it is implemented taking into consideration models used in literature for human activity recognition and bio-logging: Input Delay Neural Network (IDNN), and Support Vector Machine (SVM). For both the IDNN and the SVM the length of the input layer is defined by the length of the input window as shown in Table \ref{Seals_inputs_dim}. In case we are applying the classifier on ''windows'' of raw data, the format consists only of the second stage (the model).  
\begin{table}[t]
\tabcolsep22pt
\caption{The table lists the number of data that composed an input window in each configuration of the dataset.}
\begin{tabular}{|c c|}\hline
\hline
\textbf{Dataset}	&  \textbf{\# Input neurons}\\
\hline
\hline
Raw\_1	& 100\\
\hline
Features\_1 & 30\\
\hline
Raw\_0.4	& 40\\
\hline
Features\_0.4 & 30\\
\hline
\end{tabular}
\label{Seals_inputs_dim}
\end{table}
The IDNN model is built with one hidden layer, and one output layer. We implement two different IDNN models, with the same structure but different activation function for \textit{the input-hidden layer (function for the connection between the input layer units and the first hidden layer units} and for \textit{hidden-output layer (function for the connection between the last hidden layer units and output layer units)}. The first configuration is developed with both the input-hidden layer and the hidden-output layer activation functions  implemented using the sigmoidal function \cite{hastie2009elements}. The second configuration is developed with the input-hidden layer activation function implemented using the Radial Basis Function (RBF) and the hidden-output layer activation function implemented using the sigmoidal function. Finally in both cases the dimension of the hidden layer is selected during the model selection.

The automatic classifier for the ``stream" format classifies each time step taking into consideration the previous time steps without applying any feature extraction. To obtain this behaviour, the classifier stage is implemented with an ESN. The algorithm takes advantage from the internal memory to classify each time step taking into account the data from the previous steps. Since the model classifies each time step of the time-series in real time then each input layer is four neurons, one for each axes of the accelerometer and one for the depth value. 

\subsection{Validation schema}
\label{Validation_Schema}

\noindent In both formats of data we split the dataset randomly with ratio 70/30 where 70\% is allocated to the training set, and 30\% is allocated to the test set. For each ML model implemented in the classifier stage, we perform the model selection by 10-folds cross validation approach to select the values of the hyperparameters \cite{hastie2009elements}.

\noindent For IDNN model, the model selection is determined by varying the number of hidden units, and by three hyperparameters: the learning rate($\eta$), the momentum ($\alpha$), and the weight decay ($\lambda$). The hidden units is selected from the set $\{1,2,3,4,5,50,100\}$. It is worth noting that an high number of units corresponds to an high memory footprint to store the weights of the model. 

\noindent The $\eta$ hyperparameter is evaluated using values from set $\{0.1, 0.01, 0.001, 0.0001\}$, and $\alpha$, and $\lambda$ values are selected from set $\{0.0, 0.0001,0.001, 0.01, 0.1\}$. The training phase is performed until the mean squared error is stabilised. For IDNN this stabilisation happens in about 1000 epochs by applying the R prop learning algorithm \cite{hastie2009elements}.

\noindent The ESN model selection involves evaluating performance whilst varying the input scaling, the leaky parameter and the number of units in the reservoir. The scaling and the leaky parameter are selected respectively in sets $\{0.01- 1\}$ and $\{0.1- 1\}$. For the number of units in the reservoir, we select a value in set $\{5-100\}$. The dimension of the reservoir affects the memory footprint of the automatic classifier functions.

\noindent Finally SVM model selection is performed concerning different kernel function, and selecting the value for the C parameter. The kernel functions evaluated are: the Radial Basis Function (RBF)(with sigma value equal to 1), Linear, and Polynomial (order 3) kernels. For each kernel we consider each dataset identified for the ``windows" format. The C parameter is selected in the set $\{100, 10, 1, 0.5\}$.

\noindent To evaluate the different classifier configurations, we compute the accuracy = $\frac{TP + TN }{TP + TN + FP + FN}$, the F1score $= 2*\frac{precision + recall}{precision * recall}$ (a measure of the performance) where precision = $\frac{TP}{PP}$ and recall = $\frac{TP}{P}$ and the memory footprint. We use True Positive(TP), True Negative(TN), False Positive(FP), False Negative(FN), Predicted Positive(PP) or number of all positive results, and Positive(P) or number of positive results that should have been returned. The models are selected considering the highest performance achieved for each configuration. 

\noindent For each of the model selected using the validation schema, the final accuracy value is obtained by calculating the average accuracy across generated random splits, re-training, and then testing the accuracy of the model. The average is computed over five random splits for each configuration of the dataset. 

\noindent The memory footprint for a neural network model consists in weights of the models, or for a SVM in support vectors, plus input window or data from a single time step, and output data.

\section*{Results}
\subsection{Identification of prey handling activity in sequence}
\noindent The IDNN model and the SVM are applied over the 4 datasets described in Section \ref{Method}. Table \ref{Seals_results_IDNN} shows IDNN models selected for each dataset with the accuracy and the F1score obtained. 

\begin{table*}[t]
\tabcolsep2pt
\caption{The table lists the results obtained with model selection over the IDNN model performed with the 10-folds cross validation.}
\label{Seals_results_IDNN}
\begin{tabular}{|c | c c c c c c|}
\hline
\textbf{Dataset}& \textbf{A.Function}	&  \textbf{\#hn} & \textbf{$\eta$} &\textbf{$\lambda$} &\textbf{ACC TEST (STD)}&\textbf{F1S TEST (STD)} \\
\hline
\hline
Raw\_1	&RBF& 50&0.1&0.01&82.79\%($\pm$ 0.009)&80.18\%($\pm$ 0.009)\\
\hline
Features\_1 &RBF& 50&0.1&0.1&87.40\%($\pm$ 0.001)&89.12\%($\pm$ 0.004)\\
\hline
Raw\_0.4	&SIGM& 50&0.001&0.01&81.10\%($\pm$ 0.006)&78.41\% ($\pm$ 0.005)\\
\hline
Features\_0.4 &RBF& 5&0.01&0.0001&80.24\% ($\pm$ 0.015)&75.53\% ($\pm$ 0.016)\\ 
\hline
\end{tabular}

\end{table*}

\noindent We perform the model selection following the validation schema specified in Section \ref{Validation_Schema}. The IDNN results provide an accuracy of between $71\%$ to $81\%$ on test set with 5 to 50 neurons respectively, in the hidden layer. The increasing in accuracy obtained with sigmoidal transition function is stabilised with more than 50 neurons. A similar behaviour is with RBF transition function. In this case the accuracy obtained with 50 neurons is 82\% with test set. The model selected (see Section \ref{Validation_Schema}) for the IDNN with Raw\_1 is an IDNN with 50 hidden neurons, $\eta = 0.1$ and $\lambda = 0.01$ with the RBF transition function.

\noindent By reducing the sequence dimension, the model shows the same behaviour with 50 neurons in hidden layer. With the Raw\_0.4 dataset, the IDNN model obtains an accuracy of 71\% with 5 neurons in hidden layer. However the accuracy increases of 10\%  to 81\% with 50 neurons with the sigmoidal transition function. Changing the transition function to RBF provides an accuracy of 80\% with 50 neurons. Hence, the IDNN identified in the model selection has 50 hidden neurons, with sigmoidal transition function, $\eta = 0.001$ and $\lambda = 0.01$. 

\noindent Next we analyse the results obtained with features extracted. With the Features\_1 dataset, the accuracy obtained is 78\% and 85\% with 5 to 50 neurons respectively, in the hidden layer with sigmoidal transition function. The increase in accuracy is not significant with a number of neurons up from 50 to 100. We also observe a saturation of the accuracy with the increase of  number of neurons in the hidden layer. The same behaviour is observed with the RBF transition function. The accuracy obtained is between 78\% to 87\% respectively, with 5 to 50 neurons in the hidden layer. With more than 50 neurons we observe an increase in accuracy of less than 2\%. Hence, the selected model is an IDNN with the RBF transition function, 50 hidden neurons,  $\eta = 0.1$ and $\lambda = 0.1$. It is worth noting that compared to the accuracy obtained with the original dataset, Raw 1, the accuracy has increased by 5\%.

\noindent The accuracy obtained using the Features\_0.4 dataset is 1\% lower than the accuracy obtained with the Raw\_0.4. The accuracy provided by the IDNN with sigmoidal transition function is 77\% to 80\% with respectively 5 to 50 neurons. With the RBF transition function, 80\% in accuracy is obtained with only 5 neurons. Hence, the selected model is 5 hidden neurons, with the RBF transition function,  $\eta = 0.01$, and $\lambda = 0.0001$. 

\noindent For the Raw\_1 dataset, we select the SVM with RBF kernel and $C=10$. This SVM configuration obtains an accuracy of 83\%. By reducing the sequences dimension, we select an SVM with RBF kernel and $C=10$ that obtains an increase in accuracy by 4\% reaching an accuracy of 87\%. If we apply the SVM with $C=10$ and RBF kernel across the Features\_1 data extracted from Raw\_1 the accuracy obtained is of 87\%. The accuracy of SVM decrease under 80\% if we apply the SVM to the Feature\_0.4. In this final case, we select an SVM with RBF kernel and $C=10$.

\begin{figure} 
    \centering
    \centerline{\includegraphics[width=7cm]{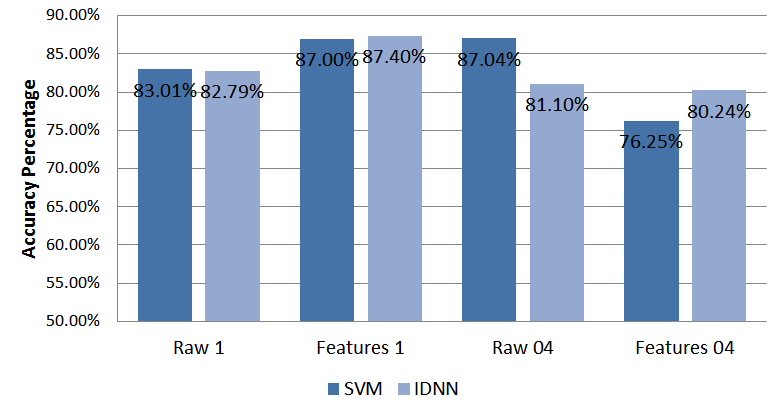}}
    \caption{Histogram summarises the results of the accuracy obtained on test set with the selected implementation of the classifier stage through IDNN and SVM.}
    \label{seals_histogram_results}
\end{figure}

\noindent The models selected over the validation set are applied over the test set. In histogram in Figure \ref{seals_histogram_results}, we show the results obtained by the selected model with the test set. We observe that we obtain similar accuracy with IDNN and SVM for both dataset formats and more important, we do not obtain an evident advantage in using features datasets for both models IDNN and SVM, then we can avoid the pre-processing phase reducing the use of computational power and memory.

\begin{table*}[t]
\tabcolsep2pt

\caption{Shown here is the memory required for each IDNN and SVM selected as well as their corresponding accuracy and F1score. The second column lists the memory required by the SVM for each selected model, and the same applies for the IDNN in the third column. For both the memory is in kilobytes. The fourth and the fifth columns give respectively the accuracy and the F1score, provided by the selected models of SVM, the sixth and the seventh columns by the selected models of IDNN.}
\label{Seals_Summarized_Results}
\begin{tabular}{|c c c c c c c|}
\hline
\textbf{Dataset} & \textbf{\begin{tabular}{c}{{}{}}SVM\\ memory\\ fp. (KB) \end{tabular}} & \textbf{\begin{tabular}{c}{{}{}}IDNN\\ memory\\ fp. (KB)\end{tabular}} & \textbf{\begin{tabular}{c}{{}{}}SVM \\acc(std)\end{tabular}} & \textbf{\begin{tabular}{c}{{}{}}SVM \\F1s(std)\end{tabular}} & \textbf{\begin{tabular}{c}{{}{}}IDNN \\acc(std)\end{tabular}} & \textbf{\begin{tabular}{c}{{}{}}IDNN \\F1s(std)\end{tabular}} \\
\hline
\hline
Raw\_1 &  5712.50 & 39.45 & \begin{tabular}{c}{{}{}}83.01\%\\($\pm$ 0.000)\end{tabular} & \begin{tabular}{c}{{}{}}80.08\%\\($\pm$ 0.001)\end{tabular} & \begin{tabular}{c}{{}{}}82.79\%\\($\pm$ 0.009)\end{tabular} & \begin{tabular}{c}{{}{}}80.18\%\\($\pm$ 0.009)\end{tabular}\\
\hline
Features\_1 &  1814.06 & 12.11 & \begin{tabular}{c}{{}{}}87.09\%\\($\pm$ 0.005)\end{tabular} & \begin{tabular}{c}{{}{}}86.08\%\\($\pm$ 0.001)\end{tabular} & \begin{tabular}{c}{{}{}}87.40\%\\($\pm$ 0.001)\end{tabular} & \begin{tabular}{c}{{}{}}89.12\%\\($\pm$ 0.004)\end{tabular}\\
\hline
Raw\_0.4 &  1985 & 16 & \begin{tabular}{c}{{}{}}87.04\%\\($\pm$ 0.004)\end{tabular} & \begin{tabular}{c}{{}{}}87.23\%\\($\pm$ 0.001)\end{tabular} & \begin{tabular}{c}{{}{}}81.10\%\\($\pm$ 0.006)\end{tabular} & \begin{tabular}{c}{{}{}}78.41\%\\($\pm$ 0.005)\end{tabular}\\
\hline
Features\_0.4 & 1945.08 & 1.21 & \begin{tabular}{c}{{}{}}76.25\%\\($\pm$ 0.006) \end{tabular}& \begin{tabular}{c}{{}{}}72.34\%\\($\pm$ 0.003) \end{tabular}& \begin{tabular}{c}{{}{}}80.24\%\\($\pm$ 0.009)\end{tabular} & \begin{tabular}{c}{{}{}}85.53\%\\($\pm$ 0.009)\end{tabular}\\
\hline
\end{tabular}

\end{table*}

\begin{figure} 
    \centering
    \includegraphics[width=7cm]{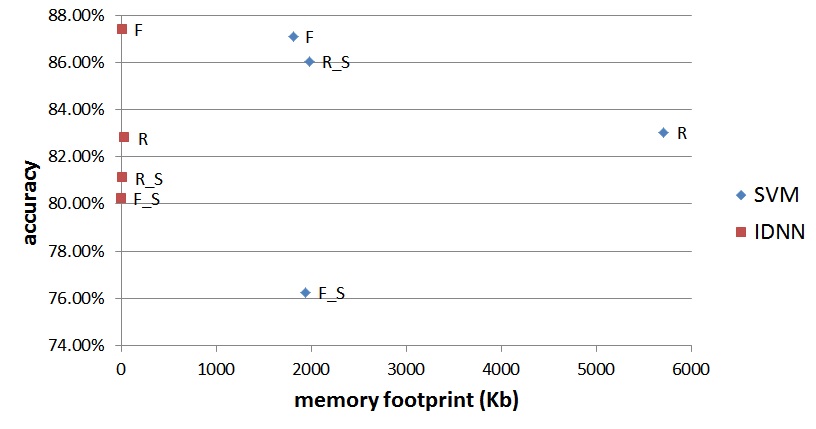}
    \caption{This graph shows the accuracy obtained for SVM and IDNN with the four dataset format and the memory footprint required for each of them.}
    \label{memoryvsaccuracy}
\end{figure}

\noindent Table \ref{Seals_Summarized_Results} shows the results obtained with IDNN and SVM across the four datasets. With Raw\_1 and Features\_1, we obtain the same performances with both the IDNN and the SVM but with a significant difference in terms of memory footprint. For the same performance the memory footprint is between 12 KB and 40 KB for the IDNN models and between 2000 KB and 6000 KB for the SVMs. For a better comprehension the results are shown in Figure \ref{memoryvsaccuracy}.

\noindent By reducing the sequences format data dimension to 0.4 (with datasets Raw\_0.4 and Features\_04), we make a similar observation about the memory footprint. The IDNN models have a memory footprint between 1 KB and 20 KB and the SVMs have a memory footprint of around 2000 KB. However, for the Raw\_0.4, the SVM has an accuracy 6\% better than IDNN. For the Features\_0.4 dataset, the IDNN accuracy is 4\% better than SVM accuracy. The trade-off result is definitively favourable to the IDNN approach.

\subsection{Identification of prey handling activity in a stream}

\noindent For the streams analysis we apply the ESN to take advantage of the internal memory inherent in this model. The ESN is applied over the raw dataset, i.e. without any filtering. The ESN analyses four values for each time step (three values from the accelerometer and one pressure value), and produces an output for each time step.

\noindent Table \ref{Seals_ESN} shows the results obtained with the ESN constructed with three different reservoir dimensions. The results indicate a stabilisation in performance with a reservoir of 5 or more units (and only a very small increase over 5 units). We obtain 86.3\% accuracy with 5, 10 and 100 units. Hence the ESN with a small reservoir of 5 units, and trained with $scaling=0.01$, $leaky=0.5$, and $connectivity=0.5\%$ is a promising solution.

\noindent Increasing the dimension of the reservoir increases the memory footprint. For example, an ESN with 5 units in the reservoir has a memory footprint of 0.4 KB whilst an ESN with 100 units the memory footprint is of 43 KB. An increase of 0.3\% in accuracy for an increase of 99\% in memory footprint. The same increase in accuracy is observed with 10 units compared to 5 units. Hence the 5 units reservoir ESN model, which achieves excellent accuracy and has a minimum memory requirement.

\begin{table}[htbp]
\setlength\tabcolsep{1pt}

\caption{The table shows the memory footprint required for each ESN model.}
\label{Seals_ESN}
\begin{tabular}{|c c c c|}
\hline
\textbf{Reservoir's \# units} & \textbf{Memory fp (KB)}&  \textbf{Accuracy Test (STD)}& \textbf{F1score Test (STD)}\\
\hline
\hline
5 &0.4& 86.3\%  ($\pm$0.0) & 85.4\% ($\pm$0.001)\\
\hline
10 &0.8&  86.3\%  ($\pm$0.0) & 86.0\% ($\pm$0.0)\\
\hline
100 &43& 86.2\% ($\pm$3.142e-05) & 85.6\% ($\pm$0.0)\\
\hline
\end{tabular}

\end{table}

\begin{figure} 
    \centering
    \includegraphics[width=7cm]{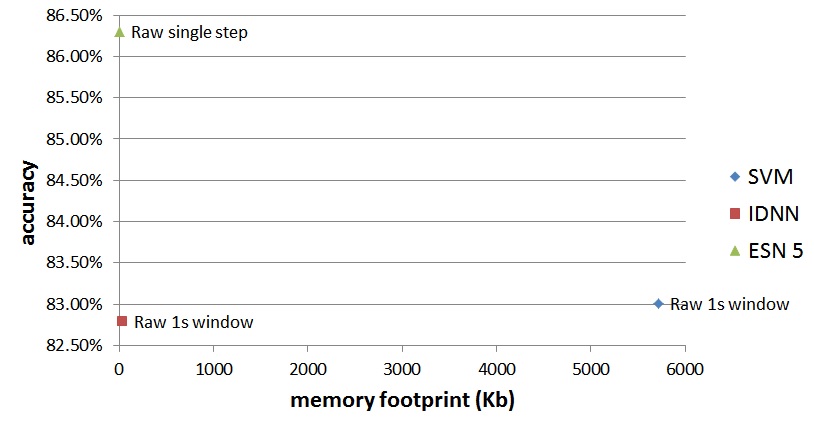}
    \caption{This graph shows the accuracy obtained for ESN compared with IDNN and SVM. The input data are raw data: with ESN the input data is a single time step, with IDNN and SVM the input data is a window of 1s.}
    \label{memoryvsaccuracyEIS}
\end{figure}

Figure \ref{memoryvsaccuracyEIS} shows the accuracy/memory footprint obtained with ESN model compared to the ones obtained with IDNN and SVM models with 1s windows. We observe that the ESN model outperforms the IDNN by $4\%$ in accuracy and 39Kb in memory footprint and the SVM model by $3\%$ in accuracy and more than 5700Kb in memory footprint. 

\subsection{Storage memory reduction for data logging}
We analyse the memory used for data storage in a bio-logging procedure to assess the advantage of using ML approaches on board. We adopt as representative of low-power devices the Galileo platform \cite{galileo}. This device is cheap and easy programmable with 8 MB of storage for the data, 256 KB RAM memory and a Intel Quark processor. Since the code is stored in a dedicated memory, for this analysis we disregard the memory required for the code. However, we compare the storage overhead for the classical bio-logging procedure, which determine how often we need to retrieve the device, with the overhead of the ML approach which only requires the activity summary to be stored and retrieved.

\noindent For bio-logging, we consider 25Hz sequences without overlap. Hence each second we have to store 100 data points, i.e. 25 data for each of the three axes of the accelerometer and 25 for the pressure/depth sensor. These data are stored as unsigned int, which occupies 4 bytes in MicaZ. This results in a storage overhead of 400 bytes per second, which fills the storage in 20971.3 seconds of data (i.e., about 6 hours). This has major implication for the entire bio-logging procedure, because it means that the researchers either have to retrieve the device almost every day in order to download the bio-logged data and free the storage or, if the device has a radio interface, the researcher needs to get within the communication range daily with the device in order to download the complete dataset.

\noindent For the ML approach, we use overlapping datasets and the cases of 1 sec and 0.4 sec sequences are considered separately. In the first case, the overlap is of 0.5 seconds. Hence the automatic classifier analyses two sequences (100 data) per second (with a delay of half second due to the overlap) and provides two (boolean) classifications per second (1 for prey handling activity and -1 if non-prey handling). Hence the storage is filled up with a rate of 2 bits per second, which takes 9320.0 hours, which is more than 389 days of activity.

\noindent Finally, if we consider the ML approach using sequences of 0.4 seconds (with an overlap of 0.2 seconds), we store 5 data of 5 bits per second. Hence, the memory is filled at a rate of 5 bits per second, and the device can operate for about 3728.2 hours (i.e. 155 days) before filling up its memory.

\noindent Table \ref{period_onboard} summarises the results obtained in this section. We observed that, by using ML approach on board, we greatly increase the autonomy of the device from 19 days up to 1 month and half. The autonomy represents the period that the device can be left on the animal, without need of retrieval and without using up the memory or losing any data.

\begin{table}[htbp]

\tabcolsep1pt
\caption{Period of autonomy of the device.}
\label{period_onboard}
\begin{tabular}{|c c|}
\hline
\textbf{Procedure} & \textbf{Period of autonomy}\\
\hline
\hline
Bio-logging  & 6 Hrs\\
\hline
ML w sequences of 1 second & 389 days\\
\hline
ML w sequences of 0.4 second &155 days\\
\hline
\end{tabular}

\end{table}


\section{Conclusions}

The aim of this paper is to propose an approach for the development of an on-board effective automatic classifier capable of identifying animal (seal prey) activities from accelerometer data. As input of the classifier, we consider the time-series data in two formats, in the first format they are split in sequences and in the second format they are streams. The automatic classifier is implemented in two stages: (i) the filter, which extracts features (which is applied only with the datasets of sequences); (ii) the classifier stage, which implemented using three different machine learning approaches.

\noindent The automatic classifier is designed to be integrated in a device worn by seals during their daily activities. In this context, we need to take into consideration the trade-off between classification accuracy and memory footprint. In particular for the classifier stage we develop and evaluate three machine learning approaches,  IDNN, ESN, and SVM, with regard to accuracy, performance and memory demands. We also distinguish differences in model performance when using sequences and streams. IDNN and SVM are employed for the classifier stage for sequences data, and ESN is employed for the classification stage for the streams data.

\noindent For the sequences formatted data, we show that, the IDNN requires significantly lower memory than the SVM whilst obtaining a similar classification accuracy. The IDNN satisfies the requirements of both high performance and generality.

\noindent In Section \ref{Results}, both the IDNN and the SVM achieve 87\% in accuracy in the filter/features extraction stage. The extraction of features in the filter stage requires a pre-processing that selects features which properly represent the sequences whilst avoiding the loss of raw information. The process of extracting features from each sequence is demanding in terms of memory capacity and time. If we take into consideration the generalisability of the automatic classifier approach to different scenarios, we observe that it is possible to avoid the ad hoc selection of features whilst maintaining high performance. We have shown that we can use the raw sequences, which are not edited by the filter stage, obtaining an accuracy of 83\% with both configuration of the classifier stage. Hence in the context of sequences, the generalisability of the IDNN model and its low memory footprint make this model highly suitable for identifying animal activities in an automatic classifier.

\noindent For the streams formatted datasets, we use an ESN model in the classification stage (without filtering). Since ESN takes advantage of an internal memory, we are able to apply the model over the stream data without pre-processing procedure. By avoiding the splitting of time-series in sequences, we obtain an higher generalisability than with the IDNN and SVM. In this context, with the term, generalisability we consider the pre-processing necessary to apply a model e.g. to apply the IDNN we need to split first the time-series in sequences, or to apply the SVM se also need to extract feature to improve results. 
ESN model analyses each time step to identify the prey handling activity and provides an accuracy of 86\% with only five units in the reservoir.

\noindent The ESN takes advantage from the entire history of the stream without the use of fixed length sequences. In in this context, the researcher is not required to decide on a sequence length, because the model analyse each time step of the stream singularly by considering the previous time steps. Here we demonstrate that the ESN obtains high performance whilst maintaining a low memory footprint. In fact, for the ESN model with five units in the reservoir only 0.4KB of memory required.

\noindent By using a machine learning approach for the classifier, much less data needs to be stored in the memory of the tag and much lower power is required. These aspects greatly reduce the challenge associated with retrieving, transmitting, and downloading time-series data associated with remote observation of species. Normally storing time-series data on board bio-logging devices, introduces the need for compression to reduce data storage needs and to reduce the energy used for transmitting this data to a remote collection station. The automatic classifier described here can be implemented on board the device and allows remote observation of the animal without storing all of the time-series data and hence avoids the need for power to transmit to a remote station. 

\noindent The results presented in this paper show that it is possible to implement efficient on-board automatic classifiers with low memory occupation and power demands, capable of supporting long duration bio-logging and animal activity recognition, under extreme environmental conditions. We expect that the automatic classifier described in this paper will provide inspiration for a new generation of Machine-Learning-based tags, that will enable ecologists to observe a wide range of animals in different environments, in real-time. These results can help to extend the application of this approach to different case studies with animals bigger then seals, such as lion seal and whales, or smaller, such as penguins. These will give a rich analysis of the application of such approach in the study of aquatic mammalian. 
\section*{Acknowledgments}

We gratefully acknowledge Monique Ladds from Macquarie University, Sydney, for the open access GitHub dataset \cite{Ladds2016seals} that we used for this research.


\bibliography{Bibliography}

\bibliographystyle{abbrv}

\end{document}